\documentclass[10pt,twocolumn,letterpaper]{article}

\usepackage[pagenumbers]{cvpr} 

\definecolor{cvprblue}{rgb}{0.21,0.49,0.74}
\usepackage[pagebackref,breaklinks,colorlinks,allcolors=cvprblue]{hyperref}

\def\paperID{33} 
\def\confName{CVPR}
\def\confYear{2025}

\title{SoccerNet-v3D: Leveraging Sports Broadcast Replays \\
for 3D Scene Understanding}

\author{Marc Guti\'errez-P\'erez and Antonio Agudo\\
Institut de Rob\`otica i Inform\`atica Industrial, CSIC-UPC, Spain
}

\begin{document}
\maketitle
\begin{abstract}
Sports video analysis is a key domain in computer vision, enabling detailed spatial understanding through multi-view correspondences. In this work, we introduce SoccerNet-v3D and ISSIA-3D, two enhanced and scalable datasets designed for 3D scene understanding in soccer broadcast analysis. These datasets extend SoccerNet-v3 and ISSIA by incorporating field-line-based camera calibration and multi-view synchronization, enabling 3D object localization through triangulation. We propose a monocular 3D ball localization task built upon the triangulation of ground-truth 2D ball annotations, along with several calibration and reprojection metrics to assess annotation quality on demand. Additionally, we present a single-image 3D ball localization method as a baseline, leveraging camera calibration and ball size priors to estimate the ball’s position from a monocular viewpoint. To further refine 2D annotations, we introduce a bounding box optimization technique that ensures alignment with the 3D scene representation. Our proposed datasets establish new benchmarks for 3D soccer scene understanding, enhancing both spatial and temporal analysis in sports analytics. Finally, we provide code to facilitate access to our annotations and the generation pipelines for the datasets\footnote{\url{https://github.com/mguti97/SoccerNet-v3D}}.
\end{abstract}

\section{Introduction}
\label{sec:intro}

Sports analytics has become an increasingly vital component in modern sports, transforming how teams, coaches, and fans understand, optimize, and interact with athletic performance~\cite{CapelleraCVPR2025,thomas2017computer, naik2022comprehensive, mendes2023survey}. The proliferation of advanced tracking technologies, such as player and ball tracking systems, has enabled the generation of rich, high-resolution data that provides unprecedented insights into the dynamics of sports competitions. The abundance of 2D tracking data has revolutionized sports analysis, allowing more informed decision-making, better player development, and the identification of strategic advantages. The ability to accurately capture and analyze tracking data now plays a vital role in sports analytics, driving advances in player performance optimization~\cite{decroos2019actions}, injury prevention~\cite{borisov2023application, blanchard2019keep}, and the development of sophisticated coaching strategies~\cite{wang2024tacticai}. While wearable tracking devices have been instrumental in generating performance data, computer vision has emerged as a compelling alternative. Employing advanced vision-based algorithms, researchers and sports organizations can extract valuable tracking data directly from video footage, eliminating the need for intrusive wearable sensors~\cite{rahimian2022optical, cui2023sportsmot}. This camera-based approach offers several advantages, such as simultaneous tracking of multiple athletes, the removal of connectivity issues associated with wearables, and the potential for retroactive analysis of historical game footage.
Computer vision-based tracking uses techniques such as object detection~\cite{cioppa2021camera}, camera calibration~\cite{agudoICPR2020,Gutierrez-Perez_2024_CVPR}, and pose estimation~\cite{yeung2024autosoccerpose} to accurately identify and monitor players, the ball, and other key elements in a sports environment. This non-invasive, data-driven approach has become increasingly sophisticated, generating rich, high-fidelity datasets that provide deeper insights into athletic performance and team dynamics.

Beyond motion tracking, computer vision has been leveraged to enhance various aspects of the sports experience. A notable example is the development of semi-automated offside detection systems~\cite{uchida2021automated, siratanita2021method}, which enable quick and accurate identification of offside positions during live matches. Similarly, automated systems for detecting player fouls~\cite{held2023vars} assist referees, improving both the fairness and pace of the game. Furthermore, real-time graphics and overlays in sports broadcasts now integrate player statistics, team formations, and tactical visualizations, enriching the viewing experience and facilitating data-driven storytelling~\cite{cavallaro2011augmenting, zollmann2019arspectator, goebert2020new, sawan2020mixed}. Looking ahead, continued advances in computer vision are set to revolutionize multiple facets of sports, from automated refereeing and advanced performance analytics to immersive fan experiences. 

While recent results in sports analytics and computer vision have unlocked unprecedented insights and experiences, a critical limitation remains: the lack of publicly available datasets that capture 3D scene information. This scarcity is primarily due to the challenges involved in recording and annotating such data, including the need for intrusive sensors, precise multi-camera synchronization, accurate calibration, and extensive manual labeling. Consequently, the development of robust 3D tracking and analysis methods is hindered by the limited availability of high-quality datasets, emphasizing the necessity for new resources to advance the field. To date, several studies have leveraged 3D information from sports videos: in basketball, multiple works have explored multi-view video processing to achieve 3D localization of players~\cite{meng2018accurate, yang20183d, monezi2020video} and ball~\cite{wu2020multi}. In~\cite{van2022deepsportradar} was introduced DeepSportradar-v1, a suite of datasets designed for tasks such as ball 3D localization, camera calibration, player instance segmentation, and player re-identification. Among these tasks, ball 3D localization truth is obtained by leveraging calibration data and annotating both the ball center and its vertical projection onto the ground. In baseball, Chiu {\em et al.}~\cite{chiu20243d} presented the MSL Baseball dataset, which captures human 3D joint positions using four synchronized cameras alongside a fixed VICON system with 10 cameras operating in tandem to track 3D poses from markers attached to the subjects. In soccer, Kazemi {\em et al.}~\cite{kazemi2013multi} introduced the KTH Multiview Football Dataset, which includes football images with 3D annotations and calibrated multi-view camera parameters. This dataset has enabled the development of various approaches for inferring 3D poses from multi-camera systems~\cite{PerezYusWACV2022}. More recently, in~\cite{jiang2024worldpose} was introduced the WorldPose dataset, which features footage from the 2022 FIFA World Cup. As a static multi-view camera setup is available, 3D human pose estimates were inferred by triangulating the 2D annotations, considering, in practice, these estimates as the ground truth. 

Despite the progress made in sports analytics and multi-view 3D understanding~\cite{meng2018accurate, yang20183d, monezi2020video, wu2020multi, van2022deepsportradar, chiu20243d, kazemi2013multi, jiang2024worldpose}, no publicly available dataset currently provides 3D ball location with ground truth in soccer, limiting advances in ball tracking and tactical analysis. To address this gap, we introduce two novel datasets denoting SoccerNet-v3D and ISSIA-3D, specifically designed to provide high-quality 3D ball localization annotations. These datasets leverage multi-view frame synchronization, accurate camera calibration, and robust triangulation to generate precise 3D ball positions. SoccerNet-v3D extends the existing SoccerNet-v3~\cite{cioppa2022scaling} dataset by incorporating refined ball annotations optimized for 3D reconstruction, offering a diverse range of camera views and ball sizes, though lacking a temporal dimension. Conversely, ISSIA-3D builds on the ISSIA~\cite{d2009semi} dataset, consisting of six static synchronized cameras, allowing for temporal 3D ball tracking but with limited camera view diversity. To enhance these datasets and establish a strong benchmarking framework, we propose a monocular 3D ball localization task. To this end, we first apply a YOLO~\cite{YourReferenceHere} object detector to locate the ball in pixel space. Then, PnLCalib~\cite{gutierrez4998149pnlcalib} is utilized to calibrate the cameras and the 3D ball position is recovered using the monocular approach in~\cite{van20223d}, setting the first baseline for the newly proposed benchmarks. The main contributions of this work are summarized as follows:
\begin{itemize}
\item We introduce SoccerNet-v3D and ISSIA-3D, the first publicly available soccer datasets with 3D ball localization annotations, leveraging multi-view synchronization and precise camera calibration.  
\item We propose a simple yet effective triangulation framework to generate 3D ball localization data from multi-view annotated datasets, providing metrics for camera calibration and triangulation to assess the quality of the generated annotations.
\item We present a bounding box optimization method that leverages ball information from multiple views to refine existing bounding boxes or generate new ones from single-point annotations, ensuring consistency within the 3D scene.
\end{itemize}

\section{Methodology}
We propose the generation of multiple-view systems built upon the combination of SoccerNet-v3~\cite{cioppa2022scaling} broadcast main-camera images along with its synchronized replay frames. Moreover, we extend the triangulation pipeline to the ISSIA~\cite{d2009semi} dataset by leveraging its multiple-view synchronized videos.

\subsection{Multiple-view systems generation}  
Given a set of synchronized images from different points of view, such as broadcast main-camera and replay frames, our goal is to perform camera calibration for each individual image using field-line annotations. Specifically, we define the input set as $\mathcal{F}=\{I^c\}_{c=1}^C$ for $C$ cameras that capture the scene simultaneously, where $c$ represents the camera index. The dataset generation process involves identifying synchronized frame groups for which camera calibration can be performed. This is achieved by estimating the projection matrices $\mathcal{P} = \{\mathbf{P}^c \in \mathbb{R}^{3\times4}\}_{c=1}^C$, ensuring $C > 1$, which allows accurate object triangulation and, therefore, 3D scene reconstruction.

\subsubsection{Framework overview}  
The first goal of the proposed framework is to achieve the localization of 3D objects through triangulation between synchronized and calibrated frames. The proposed datasets are based on the following data distributions. Firstly, SoccerNet-v3 --a major extension of the SoccerNet~\cite{giancola2018soccernet} dataset-- offers spatial annotations and cross-view correspondences for both broadcast main-camera and replay frames. It provides detailed annotations, including field lines, goal parts, players, referees, teams, salient objects, and jersey numbers, while also establishing object correspondences across different views. Secondly, the ISSIA~\cite{d2009semi} dataset, which includes ball and player annotations captured from six synchronized static cameras. The cameras are arranged in pairs, ensuring overlapping fields of view within each pair, but not across all six cameras simultaneously.

\subsubsection{Camera calibration}
We use PnLCalib~\cite{gutierrez4998149pnlcalib} as the calibration method, an optimization-based calibration pipeline that leverages a 3D soccer field model and a predefined set of keypoints. More specifically, we make use of SoccerNet-v3~\cite{cioppa2022scaling} ground-truth field-lines annotations and transform them into a geometrically derived keypoint grid that serves as input to the PnLCalib~\cite{gutierrez4998149pnlcalib} calibration pipeline.
Then, this method employs a standard full-perspective camera model as:
\begin{equation}
\label{eq:projection}
    \mathbf{P} = \mathbf{KR}[\mathbf{I}\,\vert\,\mathbf{-t}] \in \mathbb{R} ^{3\times4},
\end{equation}
where $\mathbf{R}\in\mathbb{R}^{3\times3}$ and $\mathbf{t} \in \mathbb{R}^3$ denote the extrinsic parameters (rotation and translation, respectively) to map from scene coordinates to camera ones; and $\mathbf{K}\in\mathbb{R}^{3\times3}$ denotes the camera matrix, which includes the intrinsic parameters to transform from camera coordinates to image ones. For a camera defined by a focal length $\{\alpha_x, \alpha_y\}$, a skew coefficient $s$, and a principal point $\{x_0, y_0\}$, camera matrix is defined as:
\begin{equation}
    \mathbf{K} = \begin{bmatrix}
                \alpha_x & s & x_0\\
                0 & \alpha_y & y_0\\
                0 & 0 & 1
                \end{bmatrix}	.
\end{equation}
The extrinsic and intrinsic parameters in Eq.~\eqref{eq:projection} are inferred by leveraging the coordinates of 3D object points and their corresponding 2D projections using the soccer field model as a calibration rig. Finally, the PnL optimization module enhances calibration estimates by jointly leveraging the information from detected keypoints and lines.

To evaluate the quality of the generated calibration, we make use of Magera {\em et al.}~\cite{magera2024universal} benchmarking protocol with slight modifications to adapt the metric to varying image sizes in the dataset. The evaluation relies on calculating the reprojection error between each annotated point and the line to which it belongs. Adopting a binary classification approach, each pitch segment is treated as a single entity. Therefore, a polyline representing a soccer field segment $s$ is classified as a true positive (TP) if $\forall p \in s: \min \left(d(p, \hat{s})\right) < \gamma$, being $\hat{s}$ the corresponding annotated segment, $p$ the set of points contained in the polyline, and $\gamma$ the distance threshold as a percentage of the diagonal length of the image. Otherwise, this segment is counted as a false positive (FP). Segments only present in the annotations are counted as false negatives (FN). Hence, the Jaccard index for camera calibration, $\text{JaC}_\gamma$, at a threshold $\gamma$ is defined as:
\begin{equation}
    \text{JaC}_\gamma = \frac{\text{TP}_\gamma}{\text{TP}_\gamma + \text{FN} + \text{FP}} ,
\label{eq:jac_index}
\end{equation}
where it serves as a measure of calibration accuracy and, in our case, as an indicator of the quality of the proposed annotations. 

\subsection{3D ball localization}  
\label{sec:3DBallLoc}
\begin{figure}[t!]
  \centering
  \hspace*{-0.2cm}  
 \includegraphics[scale=0.35]{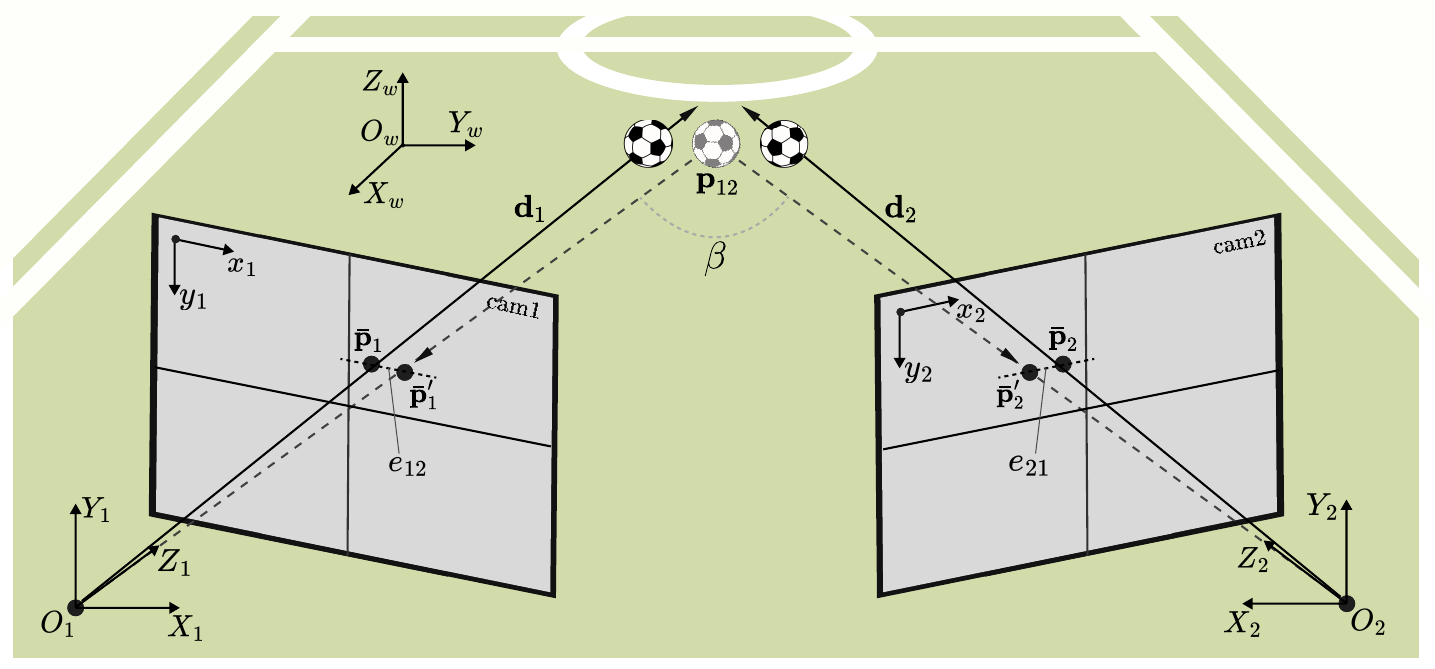}
 \caption{\textbf{Ball triangularization.} Given two cameras from different viewpoints, a 3D point $\mathbf{p}_{12}$ can be estimated from the corresponding image points $\mathbf{\bar{p}}_1$ and $\mathbf{\bar{p}}_2$ in cameras 1 and 2, respectively, using the camera projection matrices $\mathbf{P}_1$ and $\mathbf{P}_2$. However, since $\mathbf{p}_{12}$ is only an optimal solution, its reprojected image points, $\mathbf{\bar{p}}'_1$ and $\mathbf{\bar{p}}'_2$, do not exactly match the original points $\mathbf{\bar{p}_1}$ and $\mathbf{\bar{p}}_2$. $e_{12}$ and $e_{21}$ are the reprojection errors for cameras 1 and 2, respectively. $\beta$ corresponds to the parallax angle defined by the intersection of rays $\mathbf{d}_1$ and $\mathbf{d}_2$.}

\label{fig:triangulation}
\end{figure}

To estimate the 3D position of the ball, we fuse multiple 2D detections from different camera views using triangulation~\cite{hartley2003multiple}. Triangulation is the process of determining a point in 3D space given its projections onto two or more images, provided that the camera calibration matrices are known. Consider two cameras as an example, as illustrated in~\cref{fig:triangulation}. The goal of triangulation is to estimate the 3D point $\mathbf{p}_{12}$ given two corresponding 2D points, $\mathbf{\bar{p}}_1$ and $\mathbf{\bar{p}}_2$, in image coordinates, along with their respective camera projection matrices, $\mathbf{P}_1$ and $\mathbf{P}_2$. Ideally, $\mathbf{p}_{12}$ should be located at the intersection of the two projection rays, $\mathbf{d}_1=\mathbf{K}_1^{-1}\mathbf{\bar{p}}_1$ and $\mathbf{d}_2=\mathbf{K}_2^{-1}\mathbf{\bar{p}}_2$. However, these rays do not always intersect precisely due to noise introduced by lens distortion, annotation inaccuracies, and calibration errors. Instead, triangulation seeks an optimal 3D point that best fits the observed 2D detections. Given that errors in 2D ball detections and calibration annotations can lead to inaccurate triangulated positions, we use the reprojection error as a measure of triangulation accuracy. 
The reprojection error quantifies the difference between the detected 2D ball positions and the back-projected 2D positions of the estimated 3D ball. To refine the 3D localization, we filter out triangulated points with reprojection errors exceeding a predefined threshold $\tau$. The final 3D ball position is obtained by averaging the remaining valid triangulated points as:  
\begin{equation}
    \mathbf{p} = \frac{1}{N}\sum_{1\leq i, j\leq C}(\mathbf{p}_{ij}|e_{ij}<\tau),
    \label{eq:triangulation}
\end{equation}  
where $\mathbf{p}\in\mathbb{R}^{3}$ represents the estimated 3D ball position, $\mathbf{p}_{ij}$ is the triangulated 3D position from cameras $i$ and $j$, $e_{ij}$ denotes the corresponding reprojection error, $\tau$ is the reprojection error threshold, $C$ is the total number of cameras, and $N$ is the number of valid triangulated pairs with $e_{ij} < \tau$. However, due to the inherent optimality of $\mathbf{p}_{ij}$, triangulations with small parallax angles $\beta$—the angle between the two projection rays—exhibit higher uncertainty despite potentially yielding low reprojection errors. This introduces additional uncertainty in 3D ball localization, which should be considered when interpreting results.  

\subsubsection{Single-image 3D ball localization}
\label{sec:Monocular3Dball}
As a baseline for 3D ball localization from a single image, we adopt the approach proposed in \cite{van20223d}. From a monocular point of view, 3D ball localization can be performed using calibration information and knowledge of the actual ball diameter, in meters, together with its diameter and position in the image pixels space. Given the diameter $d$ and ball position $\mathbf{\bar{p}}=[\bar{p}_x, \bar{p}_y]^{\top}$ in pixels in the image space, the 3D projection rays of the ball center $\mathbf{p_c}$ and two diametrically opposed ball edges $\mathbf{p_c^+}$ and $\mathbf{p_c^-}$, expressed in the camera coordinate system, are:
\begin{align}
   &\mathbf{p_c} = \mathbf{K^{-1}} \begin{bmatrix}
        \bar{p}_x\\
        \bar{p}_y\\
        1
    \end{bmatrix} \quad ,\\
    &\mathbf{p_c^{\pm}} = \mathbf{K^{-1}} \begin{bmatrix}
        \bar{p}_x\\
        \bar{p}_y\pm \frac{d}{2}\\
        1
    \end{bmatrix}.
\label{fromdiameter}
\end{align}
Hence for a camera placed at $\mathbf{t}\in\mathbb{R}^3$ in the world coordinate system, with an orientation defined by $\mathbf{R}\in\mathbb{R}^{3\times3}$, and the true ball diameter $\phi$ in meters, the 3D ball localization is given by: 
\begin{equation}
    \mathbf{p} = \mathbf{R}^{\top} \frac{\phi\mathbf{p_c}}{\left \lVert \mathbf{p_c^{+}} - \mathbf{p_c^{-}}\right \rVert} + \mathbf{t}.
    \label{eq:ball3Destimation}
\end{equation}

\begin{figure*}[t!]
  \centering
  \includegraphics[scale=0.115]{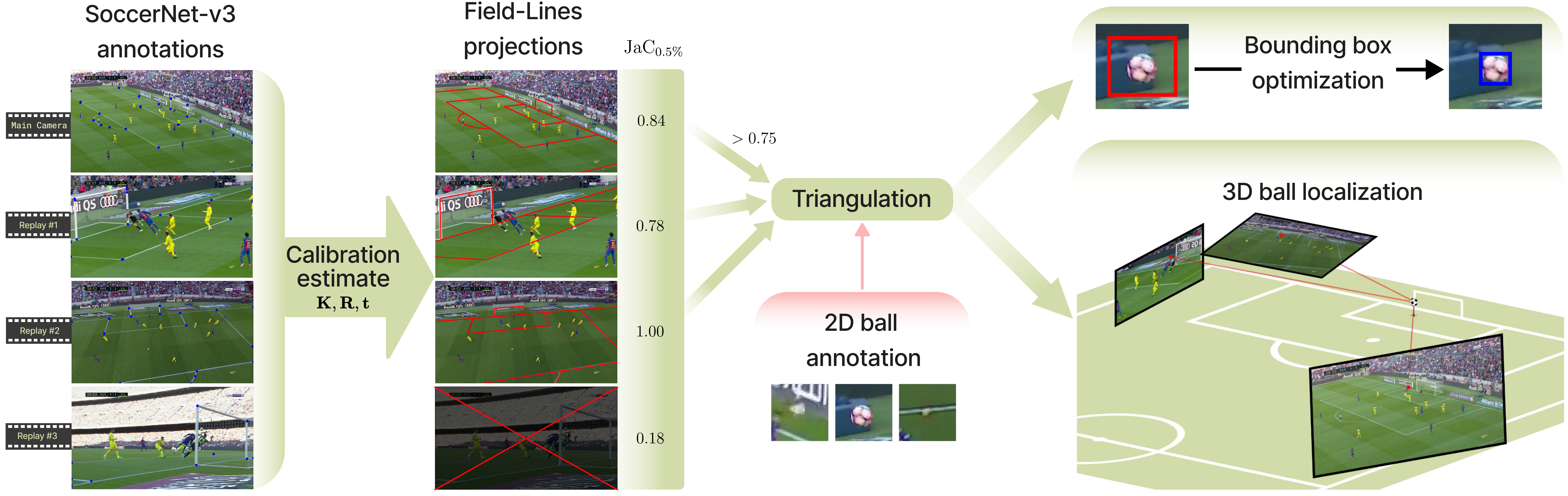} 
\caption{\textbf{SoccerNet-v3D dataset generation pipeline.} A main camera frame is paired with its corresponding synchronized replay frames, where blue dots indicate the original SoccerNet-v3~\cite{cioppa2022scaling} field-line annotations. The PnLCalib~\cite{gutierrez4998149pnlcalib} calibration pipeline is used to recover camera parameters  $\{\mathbf{K},\mathbf{R},\mathbf{t}\}$. Calibration quality is assessed using $\text{JaC}_\gamma$, with a threshold of $\text{JaC}_{0.5\%}=0.75$ to determine whether frames qualify as part of the multi-view system. Red lines represent the field projection obtained from the estimated calibration. Finally, 2D ball annotations are fused through triangulation to estimate 3D ball positions, while original bounding boxes are optimized to ensure consistency with the 3D scene, with the original SoccerNet-v3~\cite{cioppa2022scaling} and optimized bounding boxes represented in red and blue, respectively.}  
  \label{fig:pipeline}
\end{figure*}

\section{Dataset Generation}
To the best of our knowledge, in the domain of soccer, only two datasets provide 2D ball position annotations alongside multi-view camera frames:

\textbf{SoccerNet-v3}~\cite{cioppa2022scaling} is a soccer dataset comprising 33,986 images with varying resolutions from 960$\times$540 to 1920$\times$1080. It includes both main-camera and replay frames, resulting in a total of 12,764 multi-view systems. The ball position is manually annotated using bounding boxes, while the field lines are annotated by placing as many points as needed to fit them with segments formed by those points.

\textbf{Soccer-ISSIA}~\cite{d2009semi} is a two-minute soccer sequence captured using six synchronized cameras at a resolution of 1920$\times$1080 pixels and a frame rate of 25 frames per second. The ball position was annotated using a semi-automated process across all six streams, enabling 3D localization through known calibration parameters. The full field coverage is achieved through overlapping views from pairs of cameras.

In this work, we focus on soccer, where the ball appears relatively small in image space, is frequently occluded, often exhibits poor contrast, and is affected by noise and motion blur due to the sport's fast-paced dynamics. To address these challenges, we construct our datasets, SoccerNet-v3D and ISSIA-3D, extending the original SoccerNet-v3~\cite{cioppa2022scaling} and ISSIA~\cite{d2009semi} datasets, respectively.

\subsection{From SoccerNet-v3~\cite{cioppa2022scaling} to SoccerNet-v3D}
\begin{figure}[h]
  \centering
  \hspace*{-0.2cm}  
  \includegraphics[scale=0.62]{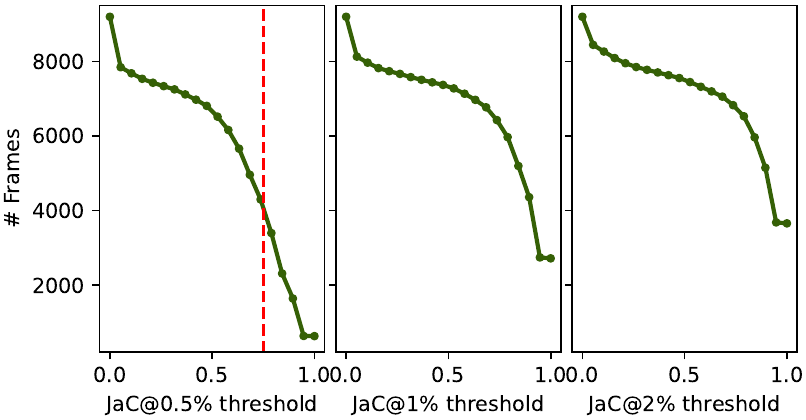} 
  \caption{\textbf{Available multi-view systems} with ball bounding box annotation on the SoccerNet-v3~\cite{cioppa2022scaling} dataset in terms of the $\text{JaC}_\gamma$ threshold for $\gamma=\{0.5,1,2\}\%$. Red dashed line represents the selected threshold corresponding to $\text{JaC}_{0.5\%} > 0.75$.}
  \label{fig:jac_index_threshold}
\end{figure}

Starting from the raw SoccerNet-v3~\cite{cioppa2022scaling} distribution, the first filtering step involves selecting the subset of images that can be calibrated using ground-truth field-line annotations. Due to the limitations of PnLCalib~\cite{gutierrez4998149pnlcalib}, a minimum number of keypoints from the predefined grid is required to obtain an initial calibration estimate. Moreover, fisheye shots from inside the goals, with extreme lens distortion, make calibration unable. While this estimate serves as a camera calibration annotation derived from field-line annotations, we further assess its quality using the $\text{JaC}_\gamma$ index, as described in~\cref{eq:jac_index}. Specifically, we compute $\text{JaC}_\gamma$ for $\gamma=\{0.5,1,2\}\%$. As the confidence threshold for the annotation increases, the number of available multi-view systems with calibration decreases, as illustrated in~\cref{fig:jac_index_threshold}. To balance annotation reliability with a sufficient number of samples, we adopt a threshold of $\text{JaC}_{0.5\%} > 0.75$, resulting in a total of 4,297 calibrated multi-view systems.

However, despite $\text{JaC}_\gamma$ filtering, certain annotation errors may persist. These errors can be categorized as follows: (1) inaccuracies in the keypoints forming the field-line annotations, (2) inconsistencies between main-camera and replay frames, where field-line annotations are correct in image space but represent different field regions due to perspective ambiguities (e.g., the main-camera annotation corresponds to the right side of the field, while the replay annotation represents the left one), and (3) discrepancies in multi-view systems where images do not correspond to the same exact moment in the match. To address these issues, erroneous images are further filtered using the reprojection error method described in~\cref{eq:triangulation}.

Additionally, enforcing the requirement that ball annotations must be present in multiple views for 3D ball localization via triangulation further reduces the number of available multi-view systems. Furthermore, as discussed in~\cref{sec:3DBallLoc}, the small parallax angle problem affects triangulation accuracy. We identify a subset of multi-view systems in which the replay view closely matches the main-camera view, leading to unreliable 3D ball localization estimates. To mitigate this issue, we impose a minimum camera displacement of 1 meter between frames. This final refinement step results in the proposed SoccerNet-v3D, the first publicly available soccer dataset with 3D ball localization annotations. SoccerNet-v3D comprises 4,051 images with precise 3D ball position annotations. An overview of the dataset generation pipeline is illustrated in~\cref{fig:pipeline}.

\subsection{Bounding box optimization}
\label{sec:bboxopt}
Although SoccerNet-v3~\cite{cioppa2022scaling} provides manually labeled ball annotations in the form of bounding boxes, we identify slight inaccuracies, particularly in the tightness of these boxes. Even if the ball is assumed to be centered within the bounding box, the estimated ball diameter in pixels using the bounding-box dimensions may be significantly erroneous. To address this issue, we employ the proposed methodology to refine the dimensions of the bounding box, ensuring consistency with the 3D representation.  

Given an image with ground-truth field-line annotations --providing a calibration estimate $\{\mathbf{K},\mathbf{R},\mathbf{t}\}$-- and a 3D ball localization $\mathbf{p}$ obtained via triangulation using~\cref{eq:triangulation}, we compute the 3D ray corresponding to the ball center using~\cref{fromdiameter}. By fixing the true ball diameter $\phi$ in meters, we optimize the bounding box dimensions through local minimization~\cite{press1989numerical} of the 3D localization error by solving the problem:
\begin{equation}
\label{min_diametre}
     \arg\min_{d} \left\| \mathbf{p} - \mathbf{R}^{\top} \frac{\phi \mathbf{p_c}}{\left \lVert \mathbf{p_c^{+}}(d) - \mathbf{p_c^{-}}(d)\right \rVert} - \mathbf{t} \right\| .
\end{equation}

After varying the ball diameter in pixels, $d$, we effectively traverse the 3D ray toward the ball center, obtaining a $d_{\text{opt}}$ in Eq.~\eqref{min_diametre} that minimizes the Euclidean distance to the 3D location label, $\mathbf{p}$. Therefore, the optimized bounding box will be centered on the 3D ray with a width and height equal to $d_{\text{opt}}$. This process ensures that the bounding boxes remain consistent with the 3D scene representation. 

\subsection{From ISSIA~\cite{d2009semi} to ISSIA-3D}
Unlike the SoccerNet-v3~\cite{cioppa2022scaling} distribution, the ISSIA~\cite{d2009semi} dataset does not include any calibration-associated annotations. To address this, and given that it consists of six static cameras, we manually annotate field-line labels following the SoccerNet-v3~\cite{cioppa2022scaling} format. This enables the generation of camera calibration annotations using the PnLCalib~\cite{gutierrez4998149pnlcalib} calibration pipeline. Notably, all six cameras satisfy the condition $\text{JaC}_{0.5\%} > 0.75$, ensuring reliable calibration quality.\\
Ball annotations in ISSIA~\cite{d2009semi} are not provided in a bounding box format but rather as a single point indicating the ball center in image coordinates. As with the SoccerNet-v3~\cite{cioppa2022scaling} dataset, 3D ball localization is obtained through triangulation using the method described in \cref{eq:triangulation}. This method is also leveraged to filter out erroneous annotations based on reprojection errors. Although the ball is never visible in all six cameras simultaneously due to their spatial distribution around the field, this approach yields the ISSIA-3D dataset, comprising 10,544 images with precise 3D ball position annotations.

Furthermore, ISSIA~\cite{d2009semi} dataset single-point ball annotations are enhanced with generated bounding boxes using the previously described bounding box optimization method. This refinement enhances the dataset's applicability for object detection tasks.

\section{Experiments}
The implementation details and metrics used to conduct this
research are presented below.

\begin{figure*}[t!]
  \centering
  \hspace*{-0.22cm}  
  \includegraphics[scale=0.13]{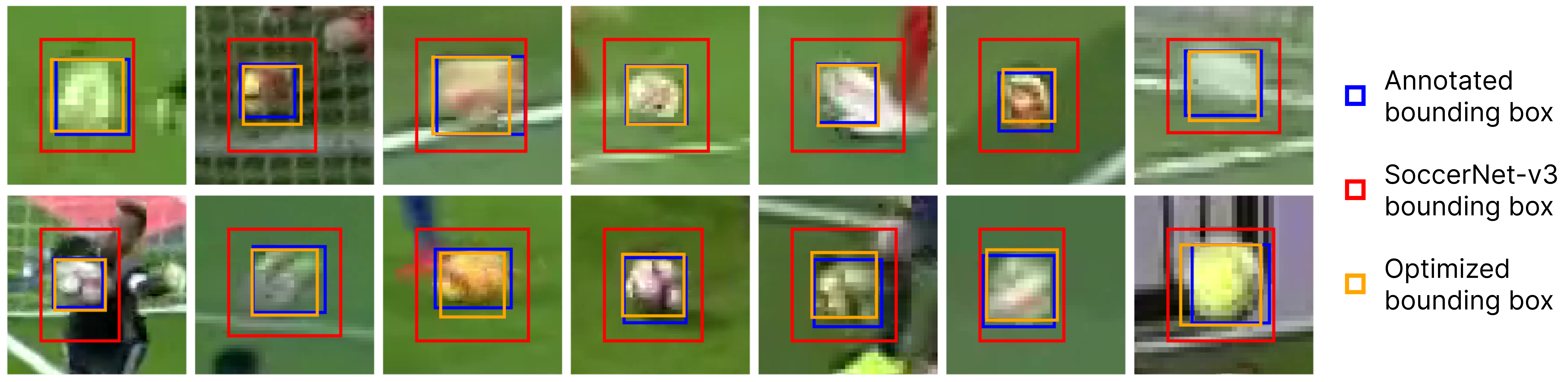} 
  \caption{\textbf{A visual comparison} of the precise manually annotated ball bounding boxes (\textcolor{blue}{blue}), the original SoccerNet-v3~\cite{cioppa2022scaling} ball bounding boxes (\textcolor{red}{red}), and the optimized bounding boxes (\textcolor{orange}{orange}) obtained through the optimization pipeline described in \cref{sec:bboxopt}.}
  \label{fig:ballans}
  \vspace*{-0.2cm}
\end{figure*}

\begin{table}[t]
\centering
\normalsize
\renewcommand{\arraystretch}{0.9}
\setlength{\tabcolsep}{5pt}
\begin{tabular}{l|cc}
\specialrule{.15em}{0em}{.5em} 
Bounding box  & IoU $\uparrow$ & Size error [\%] $\downarrow$ \\ \bottomrule \toprule
SoccerNet-v3~\cite{cioppa2022scaling}  & $\text{0.57}$ & $\text{19.01}$\\
Optimized SoccerNet-v3 & $\textbf{0.66}$ & $\textbf{7.27}$\\

\specialrule{.15em}{.5em}{.0em} 
\end{tabular}
\caption{\textbf{Bounding-boxes comparison and evaluation} against the manually annotated ball bounding boxes. Evaluation metrics include IoU and bounding box size error, expressed as a percentage of the annotated bounding box diagonal.}
\label{tab:bboxopt}

\end{table}

\subsection{Datasets}
\textbf{SoccerNet-v3D:} This dataset is built upon SoccerNet-v3~\cite{cioppa2022scaling} by leveraging both the main camera and the replay frames to calibrate the 3D scene. A total of 4,051 frames with 3D ball localization annotations are included, split into train (3,240) and test (811) sets. The dataset presents a significant challenge due to the diverse camera viewpoints, as replay frames introduce a wide range of perspectives beyond the main camera shots.  

\textbf{ISSIA-3D:} This dataset is constructed from ISSIA~\cite{d2009semi} by employing synchronized cameras for 3D scene calibration. Cameras 3 to 6 are designated as the train set, while cameras 1 and 2 form the test one, containing 8,686 and 1,858 frames, respectively. Unlike SoccerNet-v3D, ISSIA-3D includes temporal information, making it particularly suitable for additional 3D ball tracking tasks.

\subsection{Optimized bounding boxes}  
To validate the effectiveness of our optimization method, we manually annotate precise ball bounding boxes on approximately $10\%$ of the SoccerNet-v3D training set images. A quantitative comparison between the original bounding boxes and those optimized using the method described in~\cref{sec:bboxopt} is presented in~\cref{tab:bboxopt}. For evaluation, we use Intersection over Union (IoU) as the primary metric. Additionally, to account for varying image resolutions and bounding box sizes, we measure the bounding box size error as a percentage of the annotated bounding box diagonal length. Results demonstrate that the optimized bounding boxes outperform the original SoccerNet-v3~\cite{cioppa2022scaling} ball annotations on the proposed metrics, confirming their reliability as an annotation source. Furthermore, a visual comparison of the original and optimized bounding boxes is provided in~\cref{fig:ballans}, illustrating improvements in localization accuracy.

\begin{figure}[h]
  \centering
  \hspace*{-0.2cm}  
  \includegraphics[scale=0.28]{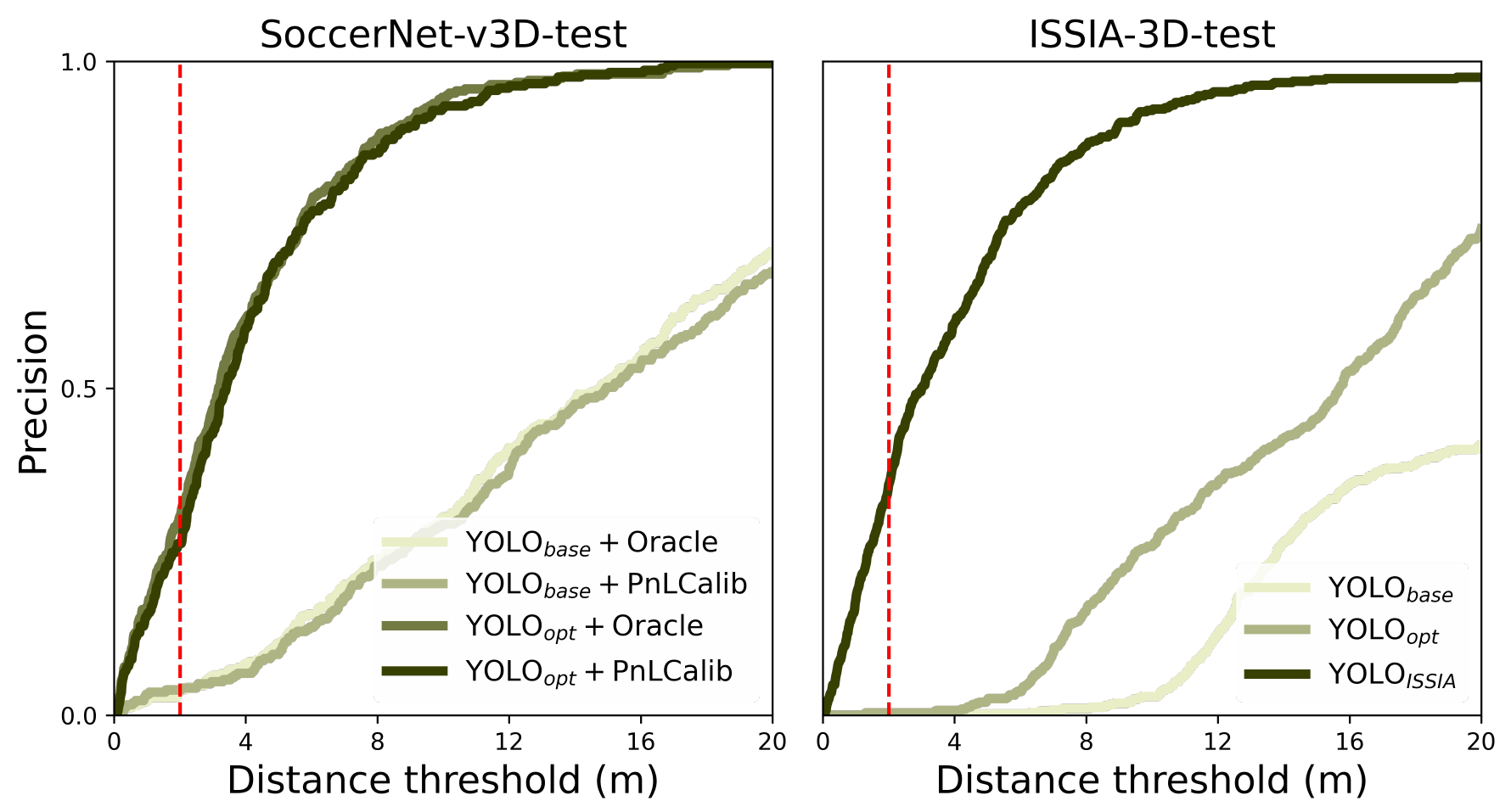} 
    \caption{\textbf{Precision-threshold curve for 3D ball localization} in SoccerNet-v3D (left) and ISSIA-3D (right). The red dashed line represents the fixed threshold $\tau_{3D}$ used for 3D localization results comparison.}  
  \label{fig:precision}
\vspace{-0.2cm}
\end{figure}

\subsection{Evaluation metrics}
We assess 2D ball detection performance using Average Precision (AP) at an IoU threshold of 0.5 (AP@0.5)~\cite{everingham2010pascal}, which serves as the standard evaluation metric for single-class object detection. For 3D localization evaluation we compute the projection error ($\text{MAE}_{\text{m}}$), defined as the Euclidean distance in meters between the actual ball position and the estimated one, using~\cref{eq:ball3Destimation}. In addition, we calculate the relative distance error ($\text{MAE}_{\%}$), which represents the ratio between the projection error and the distance to the cameras. To further analyze localization accuracy, we employ a precision plot in 3D space, illustrating the percentage of frames where the estimated ball position falls within a given threshold distance from the annotation. The position refers to the center of the ball, and the distance is measured as the Euclidean distance between the estimated and annotated ball centers. For a representative precision score, we set the threshold to $\tau_{3D} = 2$ meters for 3D localization.  
For the computation of $\text{MAE}_{\text{m}}$, $\text{MAE}_{\%}$, and precision metrics, only predictions with a non-zero IoU against the ground-truth bounding boxes are considered, ensuring that errors are evaluated only for valid detections.

\subsection{Baselines}
In this section, we present various experiments and evaluate performance in both 2D and 3D space. The proposed monocular 3D ball localization framework comprises three key tasks: 2D ball detection, field localization, and camera calibration.\\

\textbf{2D Ball Detection:} For this task, we employ the state-of-the-art object detector YOLOv11~\cite{YourReferenceHere}, specifically the YOLOv11-l model. We first train an initial YOLO model, denoted as $\text{YOLO}_{\text{base}}$, on the full SoccerNet-v3~\cite{cioppa2022scaling} training set. We then fine-tune this base detector on the optimized bounding boxes of the SoccerNet-v3D training set, resulting in the enhanced model $\text{YOLO}_{\text{opt}}$. Additionally, we fine-tune $\text{YOLO}_{\text{base}}$ on the ISSIA-3D training set using its generated bounding boxes to obtain $\text{YOLO}_{\text{ISSIA}}$. 

\textbf{Field localization:} This task involves detecting key field landmarks, such as field lines, circles, and keypoints, which serve as reference points for camera calibration. For the SoccerNet-v3D dataset, we utilize the PnLCalib~\cite{gutierrez4998149pnlcalib} keypoint detector, as it has been pre-trained on the SoccerNet-v3~\cite{cioppa2022scaling} distribution. Additionally, ground-truth field-line annotations are used as an oracle baseline for evaluation. In the case of ISSIA-3D, since the dataset consists of static cameras, manually labeled ground-truth annotations in the SoccerNet-calibration~\cite{cioppa2022scaling} format are directly used for calibration.

\textbf{Camera calibration:} We apply the PnLCalib~\cite{gutierrez4998149pnlcalib} camera calibration pipeline to estimate the intrinsic and extrinsic camera parameters from 2D-3D correspondences.

\textbf{3D Ball Localization:} We employ the method proposed by~\cite{van20223d}, as detailed in~\cref{sec:Monocular3Dball}, to estimate the 3D position of the ball from a monocular viewpoint by leveraging the relationship between the detected ball dimensions in pixels and its actual diameter in meters.

\subsection{Experiment results}

\begin{table*}[t!]
\centering
\setlength{\tabcolsep}{4pt}
\renewcommand{\arraystretch}{0.9}
\normalsize
\begin{tabular}{l|c|c|cccc}
\specialrule{.15em}{0em}{.5em} 
\text{Dataset}               & \text{2D ball detector} & \text{Field localization} & $\text{AP@0.5} \uparrow$ & $\text{MAE}_{\text{m}} \downarrow$ & $\text{MAE}_{\%} \downarrow$ & $P_{2\text{m}} \uparrow$\\ 
\bottomrule \toprule
\multirow{3}{*}{\begin{tabular}[l] {@{}l@{}}SoccerNet-v3D\end{tabular}} 
& $\text{YOLO}_{\text{base}}$ & Oracle & 0.65 & 15.3 & 18.3 & 0.03  \\
& $\text{YOLO}_{\text{opt}}$ & Oracle & $\mathbf{0.81}$ & $\mathbf{4.2}$ & $\mathbf{5.5}$ & $\mathbf{0.30}$ \\ 
\cmidrule{3-7} 
& $\text{YOLO}_{\text{base}}$ & PnLCalib \cite{gutierrez4998149pnlcalib} & 0.65 & 15.8 & 18.7 & 0.03 \\ 
& $\text{YOLO}_{\text{opt}}$ & PnLCalib~\cite{gutierrez4998149pnlcalib} & $\mathbf{0.81}$ & $\mathbf{4.2}$ & $\mathbf{5.2}$ & $\mathbf{0.26}$ \\
\bottomrule \toprule
\multirow{3}{*}{\begin{tabular}[l]{@{}l@{}}
ISSIA-3D\end{tabular}}  
& $\text{YOLO}_{\text{base}}$ & Oracle & 0.04 & 28.8 & 30.1 & 0.00 \\
& $\text{YOLO}_{\text{opt}}$ & Oracle & 0.39 & 15.6 & 16.9 & 0.00 \\
& $\text{YOLO}_{\text{ISSIA}}$ & Oracle & $\mathbf{0.65}$ & $\mathbf{4.2}$ & $\mathbf{4.8}$ &  $\mathbf{0.35}$ \\
\specialrule{.15em}{.5em}{0em} 
\end{tabular}
\caption{\textbf{Error analysis on SoccerNet-v3D and ISSIA-3D test sets}. The models $\text{YOLO}_{\text{base}}$, $\text{YOLO}_{\text{opt}}$, and $\text{YOLO}_{\text{ISSIA}}$ serve as ball detection baselines. For field localization in the SoccerNet-v3D distribution, both Oracle and PnLCalib~\cite{gutierrez4998149pnlcalib} field-landmark detection methods are evaluated, whereas only Oracle detections are used for the ISSIA-3D distribution due to its static camera setup.}  
\label{tab:results}  
\vspace*{-0.2cm}
\end{table*}

The detected ball bounding boxes, together with the camera calibration, are used to compute the 3D ball position using the method described in \cref{eq:ball3Destimation} and the known ball diameter in meters. Results for the 2D detection and 3D ball localization tasks are presented in \cref{tab:results}. 
In the SoccerNet-v3D dataset distribution, $\text{YOLO}_{\text{opt}}$ significantly outperforms $\text{YOLO}_{\text{base}}$ in the $\text{AP@}0.5$ metric. This improvement is primarily due to the optimized bounding boxes used for training $\text{YOLO}_{\text{opt}}$, which benefit from fine-tuning on a dataset with more accurately fitted bounding boxes and leading to improved detection performance. For 3D ball localization, the impact of the tighter bounding boxes is evident in the $\text{MAE}$ metrics, showing a significant reduction in localization errors, both in absolute meters and relative to camera distance. There is also a notable increase in precision for $\tau_{3D}=2\text{m}$. Since both YOLO models were trained on the SoccerNet-v3~\cite{cioppa2022scaling} distribution, these results highlight the effectiveness of the optimized bounding boxes, which not only enable $\text{YOLO}_{\text{opt}}$ to predict more precise bounding boxes but also to infer more accurate ball dimensions in pixels. Similar trends are observed when using the PnLCalib~\cite{gutierrez4998149pnlcalib} keypoint and line detection method for field localization instead of ground-truth field-line annotations. Although a decline in performance might typically be expected, the results further demonstrate the method’s robustness to small deviations in calibration estimates.\\
\begin{figure}[h]
  \centering
  \hspace*{-0.2cm}  
  \includegraphics[scale=0.29]{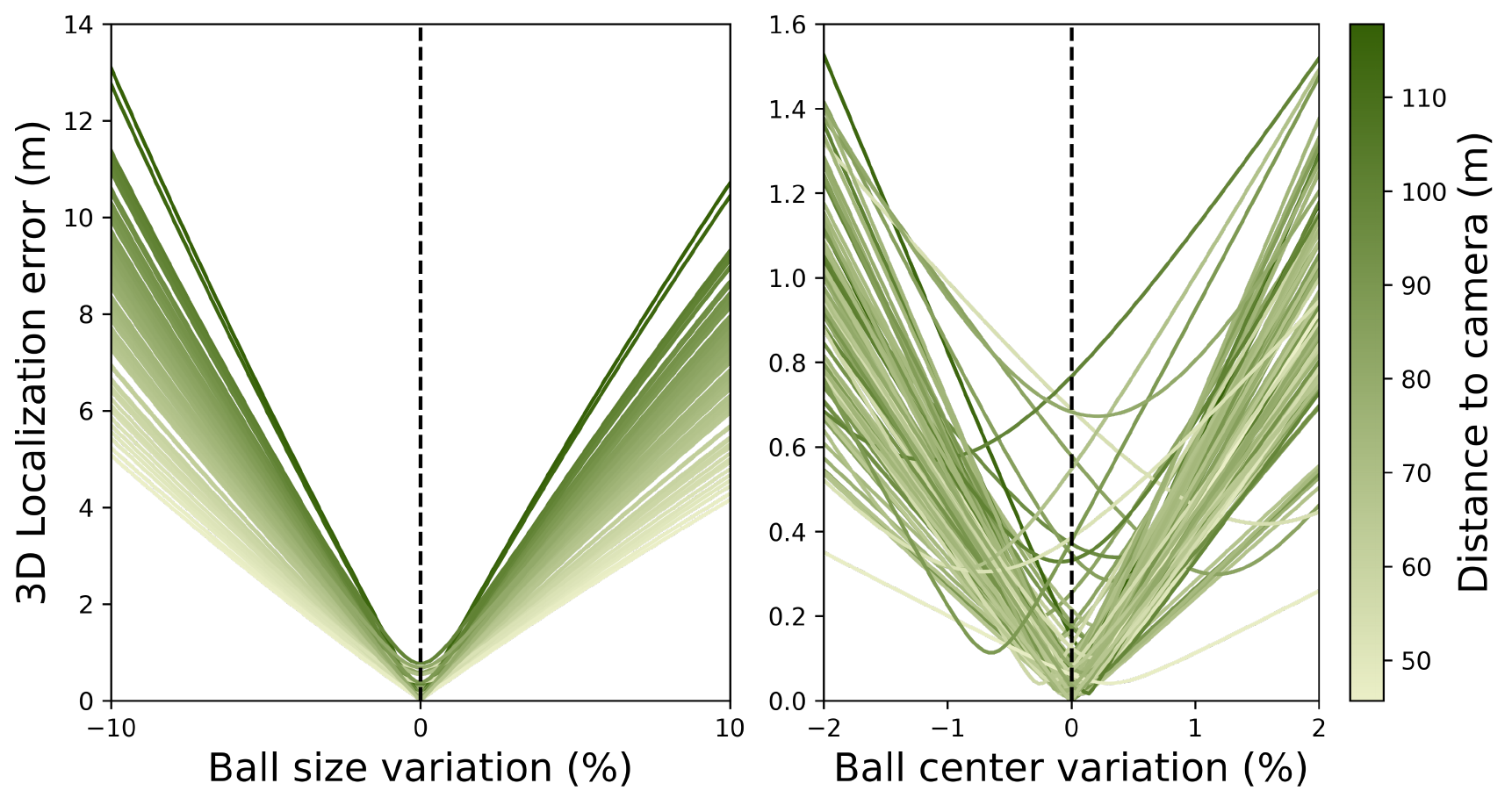} 
    \caption{\textbf{Sensitivity analysis of the 3D ball localization method} using 100 samples from the SoccerNet-v3D-train set. The left plot illustrates the 3D ball localization error as a function of ball size variation, expressed as a percentage of the bounding box size. The right plot shows the error in relation to ball center variation, measured as a percentage of the image's diagonal length. In both plots, the color scale represents the distance from the ball to the camera.}  
  \label{fig:sensitivity}
  \vspace*{-0.5cm}
\end{figure}
Results on the ISSIA-3D dataset are straightforward. YOLO models trained on SoccerNet-v3~\cite{cioppa2022scaling} data, i.e., $\text{YOLO}_{\text{base}}$ and $\text{YOLO}_{\text{opt}}$, fail to detect and localize the ball in both 2D and 3D spaces. This highlights the need for generating bounding boxes from ISSIA’s single-point ball annotations, as described in \cref{sec:bboxopt}. By fine-tuning on the ISSIA-3D generated bounding boxes, the $\text{YOLO}_{\text{ISSIA}}$ model successfully detects the ball, achieving a notable improvement in the $\text{AP@}0.5$ detection metric. Furthermore, 3D localization errors are comparable to—or even smaller than—those on the SoccerNet-v3D-test set. These results underscore the importance of the bounding box optimization pipeline, which not only allows training an object detector on a dataset without original bounding box annotations but also ensures the quality of the generated boxes.\\
The precision-threshold relationship for 3D ball localization is shown in \cref{fig:precision}, with results for the SoccerNet-v3D-test (left) and the ISSIA-3D-test (right) datasets. The former demonstrates a consistent trend with the obtained MAE metrics. Oracle field localization slightly outperforms the PnLCalib~\cite{gutierrez4998149pnlcalib} landmark detection pipeline, and $\text{YOLO}_{\text{opt}}$ achieves significantly higher precision values across all thresholds. For the ISSIA-3D-test dataset, although $\text{YOLO}_{\text{opt}}$ improves ball detection and size estimation over the base model, its generalization capability is insufficient for competitive results. This emphasizes the importance of the bounding box generation pipeline, as $\text{YOLO}_{\text{ISSIA}}$ significantly outperforms the previous models across all thresholds.\\
Finally, sensitivity analysis for the 3D ball localization method is illustrated in \cref{fig:sensitivity}, showing the behavior of 100 randomly selected samples from the SoccerNet-v3D-train set. The 3D ball localization error is evaluated by varying the annotated bounding box size and shifting the bounding box center, with error curves color scale representing the distance from the ball to the camera. The results indicate that the 3D localization error is highly sensitive to variations in the predicted ball size, ranging from approximately 6 to 14 meters for a $10\%$ pixel size variation, with sensitivity increasing as the ball’s distance from the camera grows. In contrast, sensitivity to bounding box position is less pronounced, with errors ranging from 0.6 to 1.6 meters when the bounding box position is varied by $2\%$ of the image diagonal. While some correlation with the camera distance is observed, it is notably less significant than with ball size variation.

\section{Conclusion}
In this paper, we introduce two enhanced datasets, SoccerNet-v3D and ISSIA-3D, for 3D scene understanding in soccer broadcast analysis. Leveraging field-line-based camera calibration and multi-view synchronization, we propose a monocular 3D ball localization task, which relies on triangulating ground-truth 2D ball annotations. Additionally, we present a bounding box optimization method that ensures alignment with the 3D scene representation. By combining object detectors, a camera calibration algorithm, and the real ball size in meters, we establish new benchmarks for monocular 3D ball localization task. Extensive evaluation demonstrates the feasibility of the proposed task and validates the effectiveness of the bounding box optimization algorithm. For future work, we aim to enhance our approach with more advanced monocular 3D ball localization pipelines, leverage the temporal dimension of the ISSIA-3D dataset to extend the task to 3D tracking, and explore the datasets' expansion potential for additional tasks by leveraging already-existing annotations such as player bounding boxes.

\noindent \textbf{Acknowledgment.} This work has been supported by the project GRAVATAR PID2023-151184OB-I00 funded by MCIU/AEI/10.13039/501100011033, by ERDF, UE and by the Government of Catalonia under Joan Oró FI 2024 grant.

    \small
    \bibliographystyle{ieeenat_fullname}
    \bibliography{main}

\begin{thebibliography}{40}
\providecommand{\natexlab}[1]{#1}
\providecommand{\url}[1]{\texttt{#1}}
\expandafter\ifx\csname urlstyle\endcsname\relax
  \providecommand{\doi}[1]{doi: #1}\else
  \providecommand{\doi}{doi: \begingroup \urlstyle{rm}\Url}\fi

\bibitem[Agudo(2020)]{agudoICPR2020}
Antonio Agudo.
\newblock Total estimation from {RGB} video: On-line camera self-calibration, non-rigid shape and motion.
\newblock In \emph{ICPR}, 2020.

\bibitem[Blanchard et~al.(2019)Blanchard, Skinner, Kemp, Scheirer, and Flynn]{blanchard2019keep}
Nathaniel Blanchard, Kyle Skinner, Aden Kemp, Walter Scheirer, and Patrick Flynn.
\newblock " keep me in, coach!": A computer vision perspective on assessing acl injury risk in female athletes.
\newblock In \emph{WACV}, 2019.

\bibitem[Borisov et~al.(2023)Borisov, Misnik, Velkov, and Shalukhova]{borisov2023application}
VV Borisov, AE Misnik, AA Velkov, and MA Shalukhova.
\newblock Application of computer vision technologies to reduce injuries in the athletes’ training.
\newblock In \emph{IITI}, 2023.

\bibitem[Capellera et~al.(2025)Capellera, Rubio, Ferraz, and Agudo]{CapelleraCVPR2025}
Guillem Capellera, Antonio Rubio, Luis Ferraz, and Antonio Agudo.
\newblock Unified uncertainty-aware diffusion for multi-agent trajectory modeling.
\newblock In \emph{CVPR}, 2025.

\bibitem[Cavallaro et~al.(2011)Cavallaro, Hybinette, White, and Balch]{cavallaro2011augmenting}
Rick Cavallaro, Maria Hybinette, Marvin White, and Tucker Balch.
\newblock Augmenting live broadcast sports with {3D} tracking information.
\newblock \emph{MultiMedia}, 18\penalty0 (4):\penalty0 38--47, 2011.

\bibitem[Chiu et~al.(2024)Chiu, Huang, Wu, Huang, Hsu, and Wu]{chiu20243d}
Yun-Wei Chiu, Kuei-Ting Huang, Yuh-Renn Wu, Jyh-How Huang, Wei-Li Hsu, and Pei-Yuan Wu.
\newblock {3D} baseball pitcher pose reconstruction using joint-wise volumetric triangulation and baseball customized filter system.
\newblock \emph{Access}, 12:\penalty0 117110--117125, 2024.

\bibitem[Cioppa et~al.(2021)Cioppa, Deliege, Magera, Giancola, Barnich, Ghanem, and Van~Droogenbroeck]{cioppa2021camera}
Anthony Cioppa, Adrien Deliege, Floriane Magera, Silvio Giancola, Olivier Barnich, Bernard Ghanem, and Marc Van~Droogenbroeck.
\newblock Camera calibration and player localization in soccernet-v2 and investigation of their representations for action spotting.
\newblock In \emph{CVPR}, 2021.

\bibitem[Cioppa et~al.(2022)Cioppa, Deli{\`e}ge, Giancola, Ghanem, and Van~Droogenbroeck]{cioppa2022scaling}
Anthony Cioppa, Adrien Deli{\`e}ge, Silvio Giancola, Bernard Ghanem, and Marc Van~Droogenbroeck.
\newblock Scaling up soccernet with multi-view spatial localization and re-identification.
\newblock \emph{Scientific data}, 9\penalty0 (1):\penalty0 355, 2022.

\bibitem[Cui et~al.(2023)Cui, Zeng, Zhao, Yang, Wu, and Wang]{cui2023sportsmot}
Yutao Cui, Chenkai Zeng, Xiaoyu Zhao, Yichun Yang, Gangshan Wu, and Limin Wang.
\newblock Sportsmot: A large multi-object tracking dataset in multiple sports scenes.
\newblock In \emph{ICCV}, 2023.

\bibitem[Decroos et~al.(2019)Decroos, Bransen, Van~Haaren, and Davis]{decroos2019actions}
Tom Decroos, Lotte Bransen, Jan Van~Haaren, and Jesse Davis.
\newblock Actions speak louder than goals: Valuing player actions in soccer.
\newblock In \emph{ACM SIGKDD}, 2019.

\bibitem[D'Orazio et~al.(2009)D'Orazio, Leo, Mosca, Spagnolo, and Mazzeo]{d2009semi}
Tiziana D'Orazio, Marco Leo, Nicola Mosca, Paolo Spagnolo, and Pier~Luigi Mazzeo.
\newblock A semi-automatic system for ground truth generation of soccer video sequences.
\newblock In \emph{AVSS}, 2009.

\bibitem[Everingham et~al.(2010)Everingham, Van~Gool, Williams, Winn, and Zisserman]{everingham2010pascal}
Mark Everingham, Luc Van~Gool, Christopher~KI Williams, John Winn, and Andrew Zisserman.
\newblock The pascal visual object classes (voc) challenge.
\newblock \emph{IJCV}, 88:\penalty0 303--338, 2010.

\bibitem[Giancola et~al.(2018)Giancola, Amine, Dghaily, and Ghanem]{giancola2018soccernet}
Silvio Giancola, Mohieddine Amine, Tarek Dghaily, and Bernard Ghanem.
\newblock Soccernet: A scalable dataset for action spotting in soccer videos.
\newblock In \emph{CVPRW}, 2018.

\bibitem[Goebert and Greenhalgh(2020)]{goebert2020new}
Chad Goebert and Gregory~P. Greenhalgh.
\newblock A new reality: Fan perceptions of augmented reality readiness in sport marketing.
\newblock \emph{Computers in Human Behavior}, 106:\penalty0 106231, 2020.

\bibitem[Guti{\'e}rrez-P{\'e}rez and Agudo()]{gutierrez4998149pnlcalib}
Marc Guti{\'e}rrez-P{\'e}rez and Antonio Agudo.
\newblock {PnLCalib}: Sports field registration via points and lines optimization.
\newblock \emph{Available at SSRN 4998149}.

\bibitem[Guti\'errez-P\'erez and Agudo(2024)]{Gutierrez-Perez_2024_CVPR}
Marc Guti\'errez-P\'erez and Antonio Agudo.
\newblock No bells just whistles: Sports field registration by leveraging geometric properties.
\newblock In \emph{CVPRW}, 2024.

\bibitem[Hartley and Zisserman(2003)]{hartley2003multiple}
Richard Hartley and Andrew Zisserman.
\newblock \emph{Multiple view geometry in computer vision}.
\newblock Cambridge university press, 2003.

\bibitem[Held et~al.(2023)Held, Cioppa, Giancola, Hamdi, Ghanem, and Van~Droogenbroeck]{held2023vars}
Jan Held, Anthony Cioppa, Silvio Giancola, Abdullah Hamdi, Bernard Ghanem, and Marc Van~Droogenbroeck.
\newblock Vars: Video assistant referee system for automated soccer decision making from multiple views.
\newblock In \emph{CVPR}, 2023.

\bibitem[Jiang et~al.(2024)Jiang, Billingham, Müksch, Zarate, Evans, Oswald, Pollefeys, Hilliges, Kaufmann, and Song]{jiang2024worldpose}
Tianjian Jiang, Johsan Billingham, Sebastian Müksch, Juan Zarate, Nicolas Evans, Martin Oswald, Marc Pollefeys, Otmar Hilliges, Manuel Kaufmann, and Jie Song.
\newblock Worldpose: A world cup dataset for global {3D} human pose estimation.
\newblock \emph{ECCV}, 2024.

\bibitem[Jocher et~al.(2023)Jocher, Qiu, and Chaurasia]{YourReferenceHere}
Glenn Jocher, Jing Qiu, and Ayush Chaurasia.
\newblock Ultralytics {YOLO}, 2023.

\bibitem[Kazemi et~al.(2013)Kazemi, Burenius, Azizpour, and Sullivan]{kazemi2013multi}
Vahid Kazemi, Magnus Burenius, Hossein Azizpour, and Josephine Sullivan.
\newblock Multi-view body part recognition with random forests.
\newblock In \emph{BMVC}, 2013.

\bibitem[Magera et~al.(2024)Magera, Hoyoux, Barnich, and Van~Droogenbroeck]{magera2024universal}
Floriane Magera, Thomas Hoyoux, Olivier Barnich, and Marc Van~Droogenbroeck.
\newblock A universal protocol to benchmark camera calibration for sports.
\newblock In \emph{CVPR}, 2024.

\bibitem[Mendes-Neves et~al.(2023)Mendes-Neves, Meireles, and Mendes-Moreira]{mendes2023survey}
Tiago Mendes-Neves, Lu{\'\i}s Meireles, and Jo{\~a}o Mendes-Moreira.
\newblock A survey of advanced computer vision techniques for sports.
\newblock \emph{arXiv preprint arXiv:2301.07583}, 2023.

\bibitem[Meng et~al.(2018)Meng, Xu, Li, Zeng, and Zhang]{meng2018accurate}
Weiliang Meng, Shibiao Xu, Er Li, Xiangyong Zeng, and Xiaopeng Zhang.
\newblock Accurate {3D} locating and tracking of basketball players from multiple videos.
\newblock In \emph{SIGGRAPH Asia}. 2018.

\bibitem[Monezi et~al.(2020)Monezi, Calderani~Junior, Mercadante, Duarte, and Misuta]{monezi2020video}
Lucas~Ant{\^o}nio Monezi, Anderson Calderani~Junior, Luciano~Allegretti Mercadante, Leonardo~Tomazeli Duarte, and Milton~S Misuta.
\newblock A video-based framework for automatic {3D} localization of multiple basketball players: a combinatorial optimization approach.
\newblock \emph{Frontiers in bioengineering and biotechnology}, 8:\penalty0 286, 2020.

\bibitem[Naik et~al.(2022)Naik, Hashmi, and Bokde]{naik2022comprehensive}
Banoth~Thulasya Naik, Mohammad~Farukh Hashmi, and Neeraj~Dhanraj Bokde.
\newblock A comprehensive review of computer vision in sports: Open issues, future trends and research directions.
\newblock \emph{AS}, 12\penalty0 (9):\penalty0 4429, 2022.

\bibitem[Perez-Yus and Agudo(2022)]{PerezYusWACV2022}
Alejandro Perez-Yus and Antonio Agudo.
\newblock Matching and recovering {3D} people from multiple views.
\newblock In \emph{WACV}, 2022.

\bibitem[Press et~al.(1989)Press, Flannery, Teukolsky, Vetterling, et~al.]{press1989numerical}
William~H Press, Brian~P Flannery, Saul~A Teukolsky, William~T Vetterling, et~al.
\newblock Numerical recipes, 1989.

\bibitem[Rahimian and Toka(2022)]{rahimian2022optical}
Pegah Rahimian and Laszlo Toka.
\newblock Optical tracking in team sports: A survey on player and ball tracking methods in soccer and other team sports.
\newblock \emph{Journal of Quantitative Analysis in Sports}, 18\penalty0 (1):\penalty0 35--57, 2022.

\bibitem[Sawan et~al.(2020)Sawan, Eltweri, De~Lucia, Pio Leonardo~Cavaliere, Faccia, and Roxana~Mo{\c{s}}teanu]{sawan2020mixed}
Nedal Sawan, Ahmed Eltweri, Caterina De~Lucia, Luigi Pio Leonardo~Cavaliere, Alessio Faccia, and Narcisa Roxana~Mo{\c{s}}teanu.
\newblock Mixed and augmented reality applications in the sport industry.
\newblock In \emph{EBEE}, 2020.

\bibitem[Siratanita et~al.(2021)Siratanita, Chamnongthai, and Muneyasu]{siratanita2021method}
Sirimamayvadee Siratanita, Kosin Chamnongthai, and Mitsuji Muneyasu.
\newblock A method of football-offside detection using multiple cameras for an automatic linesman assistance system.
\newblock \emph{Wireless Personal Communications}, 118\penalty0 (3):\penalty0 1883--1905, 2021.

\bibitem[Thomas et~al.(2017)Thomas, Gade, Moeslund, Carr, and Hilton]{thomas2017computer}
Graham Thomas, Rikke Gade, Thomas~B Moeslund, Peter Carr, and Adrian Hilton.
\newblock Computer vision for sports: Current applications and research topics.
\newblock \emph{CVIU}, 159:\penalty0 3--18, 2017.

\bibitem[Uchida et~al.(2021)Uchida, Scott, Shishido, and Kameda]{uchida2021automated}
Ikuma Uchida, Atom Scott, Hidehiko Shishido, and Yoshinari Kameda.
\newblock Automated offside detection by spatio-temporal analysis of football videos.
\newblock In \emph{MMSports}, 2021.

\bibitem[Van~Zandycke and De~Vleeschouwer(2022)]{van20223d}
Gabriel Van~Zandycke and Christophe De~Vleeschouwer.
\newblock {3D} ball localization from a single calibrated image.
\newblock In \emph{CVPR}, 2022.

\bibitem[Van~Zandycke et~al.(2022)Van~Zandycke, Somers, Istasse, Don, and Zambrano]{van2022deepsportradar}
Gabriel Van~Zandycke, Vladimir Somers, Maxime Istasse, Carlo~Del Don, and Davide Zambrano.
\newblock Deepsportradar-v1: Computer vision dataset for sports understanding with high quality annotations.
\newblock In \emph{MMW}, 2022.

\bibitem[Wang et~al.(2024)Wang, Veli{\v{c}}kovi{\'c}, Hennes, Toma{\v{s}}ev, Prince, Kaisers, Bachrach, Elie, Wenliang, Piccinini, et~al.]{wang2024tacticai}
Zhe Wang, Petar Veli{\v{c}}kovi{\'c}, Daniel Hennes, Nenad Toma{\v{s}}ev, Laurel Prince, Michael Kaisers, Yoram Bachrach, Romuald Elie, Li~Kevin Wenliang, Federico Piccinini, et~al.
\newblock Tacticai: an {AI} assistant for football tactics.
\newblock \emph{Nature communications}, 15\penalty0 (1):\penalty0 1906, 2024.

\bibitem[Wu et~al.(2020)Wu, Xu, Liang, Mei, and Peng]{wu2020multi}
Wanneng Wu, Min Xu, Qiaokang Liang, Li Mei, and Yu Peng.
\newblock Multi-camera {3D} ball tracking framework for sports video.
\newblock \emph{IET Image Processing}, 14\penalty0 (15):\penalty0 3751--3761, 2020.

\bibitem[Yang et~al.(2018)Yang, Xu, Wu, Zhang, and Peng]{yang20183d}
Yukun Yang, Min Xu, Wanneng Wu, Ruiheng Zhang, and Yu Peng.
\newblock {3D} multiview basketball players detection and localization based on probabilistic occupancy.
\newblock In \emph{DICTA}, 2018.

\bibitem[Yeung et~al.(2024)Yeung, Ide, and Fujii]{yeung2024autosoccerpose}
Calvin Yeung, Kenjiro Ide, and Keisuke Fujii.
\newblock Autosoccerpose: Automated {3D} posture analysis of soccer shot movements.
\newblock In \emph{CVPR}, 2024.

\bibitem[Zollmann et~al.(2019)Zollmann, Langlotz, Loos, Lo, and Baker]{zollmann2019arspectator}
Stefanie Zollmann, Tobias Langlotz, Moritz Loos, Wei~Hong Lo, and Lewis Baker.
\newblock Arspectator: Exploring augmented reality for sport events.
\newblock In \emph{SIGGRAPH Asia}. 2019.

\end{thebibliography}


\end{document}